# Metacognitive Myopia in Large Language Models

Florian Scholten, Tobias R. Rebholz, and Mandy Hütter

Psychology Department, University of Tübingen, Tübingen, Germany

*Running head:* Metacognitive myopia in LLMs

**Abstract**

Large Language Models (LLMs) exhibit potentially harmful biases that reinforce culturally embedded stereotypes, influence moral judgments, or amplify positive evaluations of majority groups. We propose metacognitive myopia as a cognitive-ecological framework accounting for a conglomerate of established and emerging LLM biases. Our theoretical framework posits that biased samples in the information environment cause five symptoms of metacognitive myopia in LLMs: integration of invalid embeddings, susceptibility to redundant information, neglect of base rates in conditional computation, decision rules based on frequency, and inappropriate higher-order statistical inference for nested data structures. Moreover, it posits that the two main components of metacognition, monitoring and control, could account for these five symptoms. Accordingly, we further outline how monitoring and control could be approximated technically, for instance, through hidden parallel reasoning histories that allow interactive LLMs to evaluate risks of myopic inference before generating overt responses. Our theoretical framework provides a novel perspective on flawed human-machine interactions and agentic AI and raises significant ethical concerns regarding the implementation of LLMs in organizational structures and high-stakes decisions.

**Metacognitive Myopia in Large Language Models**

Large Language Models (LLMs) are large-scale artificial neural networks, most commonly based on transformer architectures, that are trained on extensive text corpora to model statistical relations among tokens and generate context-sensitive natural language (Brown et al., 2020; Gemini Team et al., 2024; OpenAI et al., 2024; Vaswani et al., 2017). In this manuscript, we use the term primarily to refer to deployed generative LLM systems—such as OpenAI's ChatGPT, Google's Gemini, Anthropic's Claude, and xAI's Grok—in which pretrained models are combined with post-training, fine-tuning, safety mechanisms, or tool-use (e.g., memory components) to support open-ended interaction with users. Such systems behave in a human-like manner, primarily in the sense of generating human language for interacting with individuals across domains (Aher et al., 2023; Dillion et al., 2023; Griffin et al., 2023; Jiang et al., 2022; Rudin, 2019). They have been shown to exhibit a wide range of biases that closely resemble those long documented in human judgment and decision-making. A host of research has identified systematic biases associated with gender, race and ethnicity, religion, language, nationality, political ideology, disability, age, and other social identities (Acerbi & Stubbersfield, 2023; Bai et al., 2025; Caliskan et al., 2022; Fang et al., 2024; T. Hu et al., 2024; S. Simpson et al., 2025; A. Wang et al., 2025; D. Wang et al., 2026, 2026; Y. Yang et al., 2024; Zhao et al., 2024). That is, a broad selection of foundation models has shown ingroup favoritism and outgroup derogation (T. Hu et al., 2024), systematic disregard of variation within demographic groups and misrepresentation of minority viewpoints (A. Wang et al., 2025), or race-dependent predictions of hospitalization, costs, and health treatment recommendations (Y. Yang et al., 2024).

The prevalence of these biases is striking. Whereas heuristics-and-biases accounts (Kahneman & Tversky, 1984; Tversky & Kahneman, 1974) attribute human biases to limitations and distortions in cognitive processes, transformer LLMs with an integrated attention architecture (Gemini Team et al., 2024; OpenAI et al., 2024; Vaswani et al., 2017) are often characterized as highly effective statistical learners that capture and exploit complex regularities in large-scale data. Under optimal conditions, such as a consistently valid sample, their statistical predictions outperform human predictions (Meehl, 1954; Strachan et al., 2024). Contemporary implementations achieve near-perfect performance on various natural language processing tasks, including translation, one-shot answering of diverse questions, and the generation of sensible programming code (Brown et al., 2020). Thus, the observation that such systems, optimized for statistical prediction, reproduce many of the same biases found in humans presents an important theoretical puzzle.

A common explanation is that biased models result from the training of LLMs on biased human language corpora. Mimicking human speech also means mimicking human error, and the impression of LLMs surpassing benchmarks for natural language processing tasks was complemented by the impression of a large number of incriminating biases (Blodgett et al., 2020; Navigli et al., 2023). While this perspective undoubtedly captures some intuitively appealing truth of the phenomenon, it does not fully explain the breadth and diversity of observed biases. The claim that LLMs even precisely reproduce biases present in human language, and that such reproduction is therefore expected, has limited explanatory value for several pressing empirical and theoretical questions. While it is uncontroversial that training data reflect culturally inherent systematic social biases, this observation alone does not address which specific and systematic patterns of bias emerge in LLM outputs, how these patterns are generated, or why they persist across models and contexts. Nor does it adequately account for the emergence of apparent novel or LLM-specific biases given the models' explicit objective of learning to reproduce human language distributions (Arimanda et al., 2026). Consequently, this "language-copying" account, in isolation, contributes little to explaining the broader landscape of biases in LLMs. At best, it restates the empirical finding that biased inputs yield biased outputs, a conclusion already established in current large-scale evaluations, which consistently demonstrate that LLMs reproduce biases across a wide range of social domains (Bai et al., 2025; Kiritchenko & Mohammad, 2018; Kotek et al., 2023; Zhao et al., 2024).

However, such findings do not resolve deeper questions concerning the organization, stability, and transformation of these biases inherent to model output, nor do they clarify how these patterns should be interpreted or mitigated in practice. A more informative account of biases in LLMs must therefore move beyond the mere identification of bias, and instead engage with the structural and functional properties of LLMs that give rise to distinct and potentially categorizable biases. Consequently, the field is increasingly shifting its focus from establishing the existence of bias to identifying a conglomerate of model-inherent biases in LLMs (Navigli et al., 2023; Rich & Gureckis, 2019) in order to develop reliable methods for its measurement (e.g., Acerbi & Stubbersfield, 2023; Caliskan et al., 2022; Kocoń et al., 2023; Lin et al., 2023), benchmarking (e.g., Guha et al., 2023; S. Simpson et al., 2025; H. Yang et al., 2026), and mitigation (e.g., Christiano et al., 2017; Liu et al., 2023). However, progress in these areas is constrained by the absence of a *theoretical* framework capable of organizing and explaining a broad spectrum of observed biases. Such a framework would ideally provide (a) a theoretically informed taxonomy of biases that facilitates systematic classification, (b) explanatory principles that account for different patterns of bias within that taxonomy, and (c) guidance regarding the concepts most likely to support effective interventions.

We argue that the development of such a framework should complement attempts to engineer broadly bias-resistant systems. A psychological, cognitive-ecological approach (Fiedler, 2000a, 2012, 2014) provides a promising foundation for addressing this challenge. Rather than attributing biases in LLMs to hypothetical internal deficits of computation, it explains them in terms of the relationship between LLMs and the informational environments in which they "learned" to produce human language. Importantly, this perspective requires only minimal assumptions about the internal workings of LLMs. It treats the model as a highly effective processor of complex statistical distributions—a characterization that is broadly consistent with current evidence—while locating the origins of generated biased outputs in interactions between environmental structure (far beyond the simple language-copying account) and the absence of metacognitive monitoring and control. As such, a cognitive-ecological framework offers both a parsimonious explanation of LLM biases and a theoretically grounded basis for developing future mitigation strategies. Therefore, we propose metacognitive myopia (Fiedler, 2012; Fiedler et al., 2023) as a theoretical framework to explain potentially problematic behavior of LLMs.

**Metacognitive Myopia in Large Language Models**

In humans, metacognitive myopia is a syndrome that has been used to describe cognition as "pretty accurate in utilizing even large amounts of stimulus information, whereas [being] naïve and almost blind regarding the history and validity of the stimulus data" (Fiedler, 2012, p. 1). The theoretical framework explains biases independent of capacity limitations, lack of motivation, or depth of processing (Chaiken & Trope, 1999; Kunda, 1990). It draws heavily on the two main processes proposed in metacognition research, monitoring and control (Ackerman

& Thompson, 2017; Nelson, 1996; Thompson et al., 2011). Metacognitive monitoring deals with critically assessing all memory-, perception-, and reasoning-related cognitive processes and signaling the need of additional mental effort. Intertwined with monitoring is the process of metacognitive control, or administrating available resources and correcting invalid information. Effective monitoring and control results in effective resource allocation. Findings from diverse research domains indicate that the absence of metacognitive monitoring and control, as delineated by the metacognitive myopia framework, provides a comprehensive account of a spectrum of biases influencing human judgment and decision-making processes (Fiedler, 2012; Fiedler et al., 2002, 2019; Scholten et al., 2024; Unkelbach et al., 2007).

In contrast, LLMs are not an end in themselves but were developed for supporting humans. Particularly due to several advances in recent years, researchers continuously emphasize emerging LLM abilities that enable far-reaching universality or the capability to store and retrieve real-word knowledge (Grossmann et al., 2023; Petroni et al., 2019; Rudin, 2019). However, artificial neural networks lack the two metacognitive processes of monitoring and control. In this manuscript, we highlight that LLMs are not designed to monitor their output generation and to correct for the influence of information from erroneous sources on the basis of which statistical inferences are drawn. The integration of invalid information and erroneous higher-order statistical inferences has major downstream implications for different use cases of LLMs. For example, such flaws can affect user expectations in human-machine interactions or the implementation of domain-specific AI agents that adapt to a user's needs and thinking style. We assume that the absence of metacognition in the numerous customized applications of artificial intelligence is attributable to straightforward explanations of employed samples, prompts, and erroneous statistical inferences in the underlying computational process of LLMs.

Note that the metacognitive myopia framework most likely cannot account for the entirety of biases found in LLMs. Nevertheless, it represents a highly valuable account to develop remedies for a wide range of persistent biases in LLMs based on theoretical ideas about the influence of information distributions in the environment. By contrast, simply discovering more and more biases without such a theoretical systematization aggregates costs, and approaches to counter such biases with common methods like fine-tuning is oftentimes just fighting symptoms. We show that the metacognitive myopia framework can account for different biases of LLMs such as political bias (Feng et al., 2023), gender bias (Kotek et al., 2023; A. Xiang, 2026), and moral bias (Schramowski et al., 2022). The application of our theory facilitates the identification of remedies for influential yet biased tools used in humans' digital daily lives, beyond former explanations, such as the selection of training data and the influence of human annotators. Importantly, our theoretical framework also helps to identify the origin of novel, undocumented fallacies in, for example, the application of aggregated data to the individual level (cf. Simpson's paradox) or the equal weighting of differently reliable sources.

LLMs are entities missing metacognitive abilities. Applying metacognitive mechanisms to LLMs' computational structure aims at, first, monitoring the functioning of trained neural networks and specifically the production of text based on probability measures and, second, controlling for text products from invalid sources, invalid statistical inferences, and to-be-discarded information identified by the monitoring process. Thus, *monitoring* in LLMs reflects the system's capacity to encode and maintain representations of features that are relevant to the reliability of its own outputs. *Control*, in turn, refers to the degree to which such internal representations can be used to systematically modulate or constrain the generation process. Monitoring and control are essential for regulating the biases originating in LLMs' computational process (Johnson et al., 2026) as formulated in the cognitive-ecological sampling approach (Fiedler et al., 1996; Fiedler, 2000a), which metacognitive myopia evolved from. Sampling as a statistical metaphor refers to a decision maker just having a proportion of information (the sample; i.e., training text, subsets of information based conditionally on prompts to LLMs) about a latent distal attribute in the environment (criterion; i.e., generation of unbiased language, solution to a specific problem). In a simplified example of a sampling process, a LLM gathers information pieces according to a search algorithm from the external world or from “long-term memory” (Ma et al., 2026). From this limited observed sample, the LLM extrapolates the distribution of information within the underlying environment. As samples rarely provide an unbiased representation of the criterion, LLMs exhibit sampling biases specific to the deviations of the sample from the true latent attribute to be judged in the environment, as the sample’s implications are typically naively followed.

A key advantage of cognitive-ecological approaches is that they do not require assumptions about intrinsic psychological states or processing deficiencies to explain the emergence of biased judgments (for an extensive discussion about advantages of ecological versus intrapsychic theories, see Fiedler, 2014). These approaches conceptualize biases as the product of the informational environments from which agents sample. From this perspective, biased outputs emerge because the information available to the LLM is itself systematically structured, and the model’s responses reflect the statistical properties of the samples underlying its outputs. This framework is particularly well suited to understanding bias in LLMs. As highly effective statistical learners, LLMs can be viewed as rational agents in the ecological sense: they generate outputs that closely track the (limited) distributions of information represented in their training data and contextual inputs. Consequently, when these underlying samples contain systematic distortions, or subsamples are used based on the conditional probability of generating language given a specific prompt, the model's outputs will tend to reproduce those patterns. By precisely examining common patterns in information environments (i.e., selective, skewed, and conditional samples) that each give rise to specific biases (and not just saying that word corpora for training and fine-tuning "entail common biases of human language"), the sampling approach thus allows a systematic functional analysis of biases in LLMs. The resulting biases need not reflect failures of the model's primary information-processing mechanisms; indeed, they may arise precisely because the model accurately captures the statistical structure of the information available to it. Within this framework, the central challenge is not the generation of outputs from biased samples per se, but the absence of mechanisms capable of recognizing and correcting for such biases.

The framework of metacognitive myopia (Fiedler, 2012; Fiedler et al., 2023) evolved from the sampling approach (Fiedler et al., 1996; Fiedler, 2000a). While the sampling approach allows for analyzing behavior based on the environment (i.e., where biases originate), metacognitive myopia highlights the two metacognitive functions of monitoring and control, which are essentially designed to correct biases in decision-making (i.e., how to counteract biases). Metacognitive myopia thus describes precisely how monitoring and control in LLMs should individually address various biases through sampling within the information environment. By enabling agents to evaluate the adequacy of their own outputs and regulate responses that merely reflect local informational distributions, metacognitive processes may provide a means of counteracting biases that emerge from otherwise effective "first-order" processing. In this sense, metacognition constitutes a complementary layer of processes that can intervene when accurate representation of sampled information conflicts with broader normative or social goals (Scholten et al., 2026).

LLMs are engineered to sample a specific proportion of information with the objective of attaining the capacity to generate context-sensitive language. For example, in the process of tokenization, training text is parsed into the smallest meaningful, granular units, such as words, symbols, or bytes (Mielke et al., 2021). The resulting substrings are called tokens. The information contained in the tokenized training text can be converted to embeddings, which are number vectors that encode a token's representation within a text (for an extensive description of the functioning of LLMs, see Hussain et al., 2023). Each data point in the training data, and each human feedback during post-training which are the result of the training process and what the underlying neural network technically uses to generate new output. Similarly, any input that is not used cannot alter the model's weights. Therefore, the prevailing assumption underlying information processing is the veracity of the information. Consequently, the processing of information during training and/or production (i.e., prompts) is tantamount to categorizing it as true. As a result of being trained on large-scale corpora without distinguishing valid from invalid information, LLMs tend to exhibit a truth bias (as established for humans, Gilbert et al., 1990; Levine et al., 1999), treating encountered patterns as reliable. Hence, metacognitive myopia is most conspicuous when the falsity of a token is obvious, and so is the lack of trustworthiness of the source. Somewhat ironically, LLMs inevitably make up nonexistent sources, so-called *artificial hallucinations* (Alkaissi & McFarlane, 2023; Xu et al., 2024).

## A Symptomatology of Myopic Large Language Models

Whereas people make a choice after screening differently implicational options based on knowledge about them (Beach, 1993), LLMs decide on the next word depending on probabilities derived from token associations. These associations are calculated based on enormous data sets that LLMs were fed during training (Brown et al., 2020). LLM architectures are undoubtedly well-calibrated in the detection of correlational patterns in large web corpora and sets of books but fail to critically assess the information's origin by design. That is, LLMs not only neglect data quality (Smith, 2018), but also data genesis, including generation, context, and source validity. Therefore, LLMs show a naïve and confident reliance on every information they are trained on without questioning the source. We identify five symptoms of metacognitive myopia in LLMs (*Symptoms A – E*, see *Figure 1*), constructing a taxonomy ordered from more blatant to more sophisticated cases of metacognitively myopic LLM decision making (Fiedler, 2012). Our aim is to classify the origins of biases proposed in the LLM literature, to identify fallacies that so far have not received attention, and to illustrate the role of monitoring and control as means to solve these issues.

Note that each symptom's prevalence is likely to vary across distinct developments of LLMs with various goals and functions. The symptoms reflect broad tendencies of a broad spectrum of contemporary LLMs as a class rather than specific predictions for individual models. We further recognize that factors such as model architecture, training data composition, and alignment procedures contribute substantially to observed variation in biases across LLMs. However, we do not contend that these factors necessarily eliminate the underlying phenomena discussed here; rather, they may only influence the extent to which such phenomena manifest in particular models. Broad recent evidence further indicates that biases—viewed here as manifestations of the five proposed symptoms—are consistently observed, and often with substantial effect sizes, across a diverse set of LLMs and languages selected for investigation on the basis of their prominence and widespread deployment (T. Hu et al., 2025; Nadeem et al., 2021; Y. Yang et al., 2024; Zhao et al., 2024).

Similarly, note that the symptoms included in the present framework do not all rest on the same degree of empirical support. The initial Symptoms A-C correspond to biases that have been repeatedly documented across studies and continue to be observed in successive model generations (e.g., Ma et al., 2026; Xiang, 2026). These symptoms can therefore be regarded as comparatively well-established features of contemporary LLM behavior. In contrast, Symptoms D and E represent more exploratory predictions derived from the theoretical framework and have yet to undergo extensive empirical validation. These proposed symptoms are highlighted by an orange dotted line in *Figure 1* to distinguish them from more firmly empirically established phenomena. This distinction underscores an important strength of our metacognitive myopia framework. Its value lies not only in providing a systematic taxonomy of existing biases and identifying theoretical concepts that may support their mitigation, but also in generating novel and testable predictions. By specifying the ecological and metacognitive conditions under which particular biases should emerge, the framework offers a principled basis for anticipating previously unobserved phenomena and guiding future empirical research. Consequently, the framework serves both an organizational and an explanatory function, while also contributing to theory development through the prediction of bias patterns that have not yet been identified or systematically investigated.

**Figure 1. Symptomatology of metacognitive myopia in LLMs.**

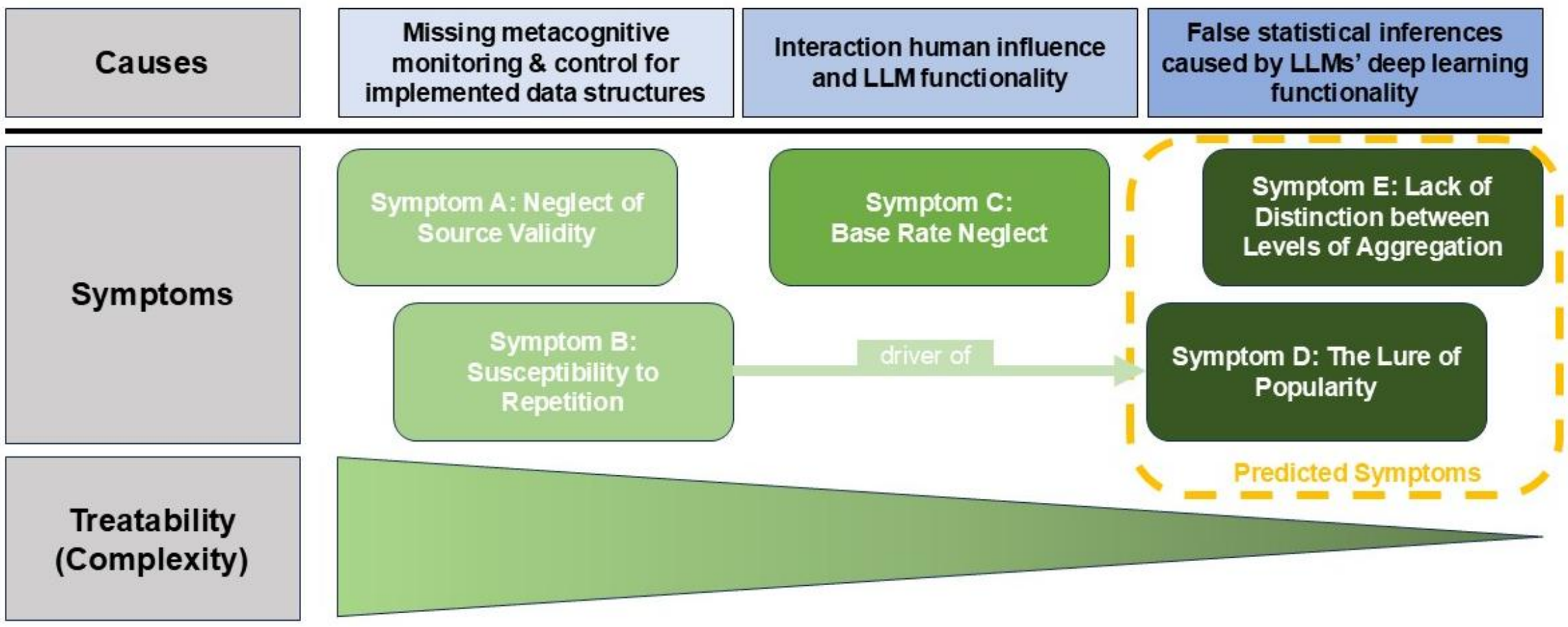

*Note.* The five symptoms of metacognitive myopia in LLMs have their origin in different LLM functionalities. Symptoms A and B are more closely related to the lack of monitoring and control processes for the mechanistic use of the underlying training data. Symptom C is an implication of the creation of a probabilistic prior through human users' prompts rendering text generation a conditionally probabilistic process. Symptoms D and E—novelly predicted from our framework as indicated by the orange dotted line—are attributable to errors resulting from false statistical inferences in correlational pattern detection. The symptoms are ordered from left to right by complexity. The more challenging a symptom is to detect through monitoring and control, and the more difficult it is to treat through engineering, the more complex the symptom becomes.

### *Symptom A: Neglect of Source Validity*

A general law of human and non-human learning dictates that when a conditioned stimulus (CS; e.g., switching on a light bulb) is continuously paired with an unconditioned stimulus (US; electric shock), the CS itself will come to elicit US reactions (e.g. fear caused by switching on a light bulb; Rescorla & Wagner, 1972). Importantly, a definition of learning as "changes in the way a system behaves towards a stimulus that are the result of regularities in the environment of that system" (De Houwer & Hughes, 2022, p. 651) extends broadly to other systems and generates predictive value for LLMs alike. For instance, each training data set token and its embeddings influence a LLM's associated parameters (Stokel-Walker & Van Noorden, 2023). Therefore, the most easily detectable symptom of metacognitive myopia, *Symptom A*, is *LLMs' inability to ignore invalid sources*. Coined mere considering (Fiedler, 2000b; Fiedler et al., 1996) or misinformation effect (Loftus, 2005) in the psychological literature, merely thinking about a stop sign (e.g., "did you see a stop sign while driving to Peter's house?") can inflict false memories. The same way the statement's subjective truth value increases through considering the possibility of the sign's presence, the probability of regenerating invalid information by LLMs increases through stronger weights promoting a specific word constellation.

The use of modern information ecologies in the training of LLMs results in the inadvertent incorporation of misinformation (e.g., fake news, hate speech) from a variety of online sources and corpora (B. Hu et al., 2024; Wulczyn et al., 2017). LLMs can reproduce such misinformation in two ways: First, simply by memorizing and citing training data (Nasr et al., 2023; Petroni et al., 2019). Second, misinformation permanently affects token embeddings, especially in cases where new language is created by combining different contexts, as in transformer architectures (Vaswani et al., 2017). In a bottom-up approach, researchers try to recover LLMs from biased output through standard practices such as reinforcement learning (Christiano et al., 2017; Y. Liu et al., 2023). Reinforcement learning is an attempt to correct for the influence of misinformation on model weights by having human annotators rate enormous amounts of LLM output as acceptable or unacceptable. This external control is problematic because human annotators are themselves biased (Sap et al., 2022). Accordingly, with reinforcement learning developers attempt to reduce the direct reproduction of biases in LLM output, but fail to recognize the lack of internal monitoring and control of invalid text pieces. This confounding of cause and effect particularly stresses the need for metacognitive myopia as a theoretical framework for biased LLMs.

***Symptom B: Susceptibility to Repetition***

*Symptom B* revolves around *LLMs' myopic processing of samples containing selectively repeated information*. Much of LLMs' information ecology is comprised by social media and Wikipedia articles by training. As web scraping can be implemented into the latest versions of LLMs (Ahluwalia & Wani, 2024), the same published journal article, opinion piece, or blog entry can be included several times in the sample used by LLMs. The pre-processing of training text by developers involves a sophisticated step of removing repetitions, filtering out toxic content, and generating a more equitable representation of languages. The biases inherent in this curated selection persist in the output of LLMs. This phenomenon can be attributed not only to developers' own biases during the sample creation process, but more crucially, to the absence of metacognitive regulation in LLMs concerning the impact of repetition on judgments and next-token prediction.

A lack of metacognitive correction pertains to both forms of redundancies, namely the training on identical sources (e.g., the same training example in another corpus), and training on the close reproduction of a text in a similar form (e.g., plagiarized, newly formulated news on another web-scraped website). LLMs fail to control for myopic sampling in that they are susceptible to redundancies of tokens and of token co-occurrences. This redundancy is likely to increase as LLMs move from being web-scraping consumers to co-writing (or even solo-writing) agents that produce an increasing proportion of the articles from which they subsequently learn (Shumailov et al., 2024). Already only 33% of news articles on the internet can be identified as original content (Cagé et al., 2020). As a consequence, the model may repeatedly endorse the same information without any monitoring of parameters, such as time of publication, uniqueness, or resumption. For example, a chorus of outdated and repeated recommendations on safety behavior during a pandemic can superimpose current recommendations, which are rarely reported due to less urgency and less public attention.

Although a simple repetition only adds informational value incrementally, the more duplicate tokens and accompanying contexts are prevalent in a training data set, the more likely is the identification of the co-occurrence, and in consequence, its reproduction. This property opens the door to a positive conception of LLMs as factual knowledge bases (e.g., in an educational context; Petroni et al., 2019) only for an unbiased sample. However, unbiased samples are rare and often dependent on a variety of auxiliary assumptions. Much like repeated and coherent information activating references in human memory (Unkelbach, Koch, Silva, et al., 2019), the repetition of a word increases its model weight and maximizes the generation probability in different contexts. Borrowing an analogy from Ludwig Wittgenstein, in such instances, LLMs function "as if someone were to buy several copies of the morning newspaper to assure himself that what it said was true" (Wittgenstein, 1977).

Interpreting the sources of training text corpora as separate agents determining the output (Binz & Schulz, 2023), LLMs can be conceived of as decision-making *groups*. As such, they can suffer from biases of selectively repeated or neglected information between corpora or 'group members.' A fallacy of collective wisdom like the hidden-profile problem (established for humans by Schulz-Hardt et al., 2006; Stasser & Titus, 1985) can illustrate the maladaptive consequences of LLMs' missing metacognitive control for repetition. Applied to LLMs' decision rules, the hidden-profile problem is illustrated by a scenario in which the LLM's choice for an ambiguous word generation task would be more accurate than each corpus' choice if only unique arguments were counted (see *Table 1* for a hypothetical exemplary distribution).[1]

**Table 1: The hidden-profile problem in LLMs.**

| | Corpus 1 | Corpus 2 | Corpus 3 |
|---|---|---|---|
| Shared Arguments | A1+, A2+, A3+, A4+, A1–, A2– | A1+, A2+, A3+, A4+, A1–, A2– | A1+, A2+, A3+, A4+, A1–, A2– |
| | B1+, B2+, B1–, B2–, B3–, B4– | B1+, B2+, B1–, B2–, B3–, B4– | B1+, B2+, B1–, B2–, B3–, B4– |
| Unshared Arguments | A3–, A4– | A5– | A6– |
| | B3+ | B4+, B5+ | B6+ |

*Note.* The table depicts a biased hypothetical distribution of shared and unshared arguments inducing a hidden-profile problem in LLMs as decision making groups. A = argument about increasing military expenditure; B = argument about decreasing military expenditure; + = argument for A/B; – = argument against A/B. Consider a prompt asking whether a government should (A) increase or (B) decrease the military expenditure. In total, three individual corpora shaped model weights in the training phase with relevant information. Together, these yield six identical arguments for decreasing (B+) the expenditure and four for increasing (A+) it, encompassing the highly similar embeddings. Conversely, together they provide four identical arguments against decreasing (B–) the expenditure and six against increasing (A–) it. In the event that all corpora share the four arguments for A and only two against A, as well as the four arguments against B and only two arguments for B, while additional arguments against A and for

[1] Note that when "arguments" or "evaluative information" are described in this article as the basis for a LLM's decision, this primarily refers to constellations of tokens in their respective context reflected in embeddings.

B are included in individual corpora only, the LLM should be more likely to generate a position favoring A after training. This is despite the overall favoring of B for the individual consideration of all arguments across corpora when simply summarizing the information environment (i.e., four arguments for A and six arguments against A resulting in a net of two arguments against A plus six arguments for B and four arguments against B resulting a net of two arguments for B). The inclusion of the exact same argument in two or more corpora does not increase the overall number of arguments, but it still positively influences the probability of reproducing the argument and, more generally, the probability of generating words that favor the intended choice (Shiflett, 1979). Note that it is inevitable that general-purpose LLMs will integrate other corpora that do not contain information about the target question yet still influence the model weights. For the sake of simplicity of this example it is assumed that the associative influences of the other corpora on the semantically independent part of the text generation is marginal.

Selective repetition is the common denominator for three further instances of prominent biases embodying culturally and historically grown predispositions in LLMs. First, word embeddings in LLMs are three to six times more likely to associate the pronoun *he* than the pronoun *she* with high-profile professions such as doctor, even overshooting the actual asymmetry observed in real-world employment statistics (Bolukbasi et al., 2016; Kotek et al., 2023). Second, LLMs reliably ascribe higher sentiment scores to sentences featuring a specific gender or race. That is, LLMs adhere to the stereotypes of females being more emotional than males and amplify stereotypes by associating African American names with anger, fear, and sadness more than European American names (see An et al., 2024; Xiang, 2026; Zhao et al., 2024, for several other instantiations of gender biases). Third, with the Word-Embedding Factual

Association Test as a modified version of the Implicit Association Test (IAT), Caliskan et al. (2017) replicated eight common IAT findings by training a simple LLM on a large-scale internet dataset. This demonstrates the transfer of biases from peoples' thinking to their written language. In turn, LLMs maintain and even facilitate these biases in the form of inflated statistical regularities inherent in the output of word embedding tasks and language production (Caliskan et al., 2022; Glickman & Sharot, 2025; Guo & Caliskan, 2021). Consequently, frequent tokens and embeddings shape the political, moral, and interpersonal biases implemented in LLMs (Kotek et al., 2023; Schramowski et al., 2022; Tan & Celis, 2019; Waight et al., 2026).

Moreover, the common structure of information ecologies predicts a positive evaluation bias. Given the differential distributions of positive and negative information in the natural human information ecology (Unkelbach et al., 2008; Unkelbach, Koch, & Alves, 2019), LLMs may reproduce and potentially reinforce positivity biases in favor of the subject at hand. A positive evaluation bias in LLMs is then entrenched in the law of repetition, whereby similar positive information is bundled and draws a more coherent picture than negative information (Alves et al., 2017; Unkelbach et al., 2008). For example, wealthy people are often perceived in a dualistic manner by the general public. Given that positive information (e.g., donations to their hometown, appearance at a celebrity gala, promotion of the arts) is more similar than negative information (e.g., difficult working conditions in factories, dubious opinions about democratic institutions, tax evasion; Unkelbach et al., 2008), positive information should have a higher coherence of embeddings in LLMs. Consequently, the generation process of LLMs could rely on a skewed distribution favoring coherent positive information about ambiguous and potentially immoral individuals or concepts, thereby obscuring the extent and nature of negative information.

### ***Symptom C: Base Rate Neglect***

In between blatant and more complex fallacies, *Symptom C* describes *LLMs' flawed conditional reasoning that rightfully relies on (at first glance) unbiased samples, but misses conditional dependencies in the sampling process*. This symptom may be attributed to the learning mechanism of choice-contingent availability inherent to the underlying network model itself or to the fact that the LLM's human interaction partner formulates requests to the models in prompts that artificially restrict the context on which an output is generated. In both cases, this yields samples that are suitable for generating output. However, due to the imposed conditionals, these samples only represent a portion of the total sample. To illustrate such base rate neglect (as in humans, Pennycook et al., 2022), one might consider the hypothetical generation of recommendations regarding music genres and artists by LLMs. In the case of a prompt requesting "popular music artists," it can be assumed that the sample of tokens and embeddings relevant to the generated output will be tailored to name pop rather than rock music artists, given that the "pop" token in the keyword "popular" is more likely to be associated with pop stars than rock stars.

LLMs trained on data from community websites like Stack Overflow or GitHub (Raffel et al., 2020) have become proficient in generating programming code due to their exposure to numerous expert responses. These forums often feature correct solutions from expert users addressing their peers' queries. However, for rare or uncommon queries, correct answers are less likely to be reproduced or generated by combining known solutions, due to the limited examples available. While LLMs can produce helpful code solutions by leveraging thousands of similar training examples, they can also generate misleading or incorrect answers, such as fabricating nonexistent R-packages or functions. For complex or novel programming tasks, LLMs rarely acknowledge their limitations, as they rely on the conditional probability of providing an answer based on the presence of expert responses in the training data, P(answer x|answer given). They neglect the conditional probability of responding with “I don’t know” based on the absence of answers from other users who faced the same query, P(answer “I don’t know”|no answer given). For highly complex tasks, the ideal response pool should include both uncertain answers and indications of no solution. However, LLMs exhibit choice-contingent availability, generating text solely based on the training data, and disregarding the importance of uncertain or missing answers (Smith, 2018).

People rarely sample unconditionally (Fiedler, 2008). Even if a user prompts without a strong conditional, training data are likely to not display the true base rates of arguments in favor of different options. With every change in the sample’s population, it would be necessary to either reset the sample or add another one (Fiedler, 2012). Consequently, unconditional sampling carries the risk of conserving LLMs pre-trained base rates that could be incongruent with the true frequencies of occurrence for latent quantities (just like humans, Bolukbasi et al., 2016; Fiedler, 2012). However, because people generally tend to sample based on underlying default attitudes, conditional sampling poses a greater danger. This is illustrated by its most extreme case, the ultimate sampling dilemma (as proposed by Fiedler, 2008) in *Figure 2*. Based on the distributional assumption of skewed frequencies, a variety of polarizing issues could produce detrimental results in LLMs depending on the geared hypothesis that creates a positive or negative evaluative context.

Finally, based on the calculation of the most probable output for a prompt, sequential interactions represent the process of Bayesian belief updating, featuring conditional probabilities for multiple prompts. Due to a continuous influence of all prompts in the interaction history, previous token vectors leave accountable traces for newly computed output. In this case, the preceding prompt-output combination(s) plus the hypothesis put forward by the new prompt set a distributional prior for the new output. By including personal opinions as context in prompts, a correspondence bias is likely to appear in LLMs (similarly to the one shown for humans; Gilbert & Malone, 1995; for LLMs, see Salvi et al., 2025). This fallacy entails that a person will infer inherent LLM attitudes from implicitly self-navigated output without metacognitively questioning one's own influence on the result (Rebholz, 2026; Rebholz et al., 2024).

**Figure 2. Information ecology for the ultimate sampling dilemma in LLMs.**

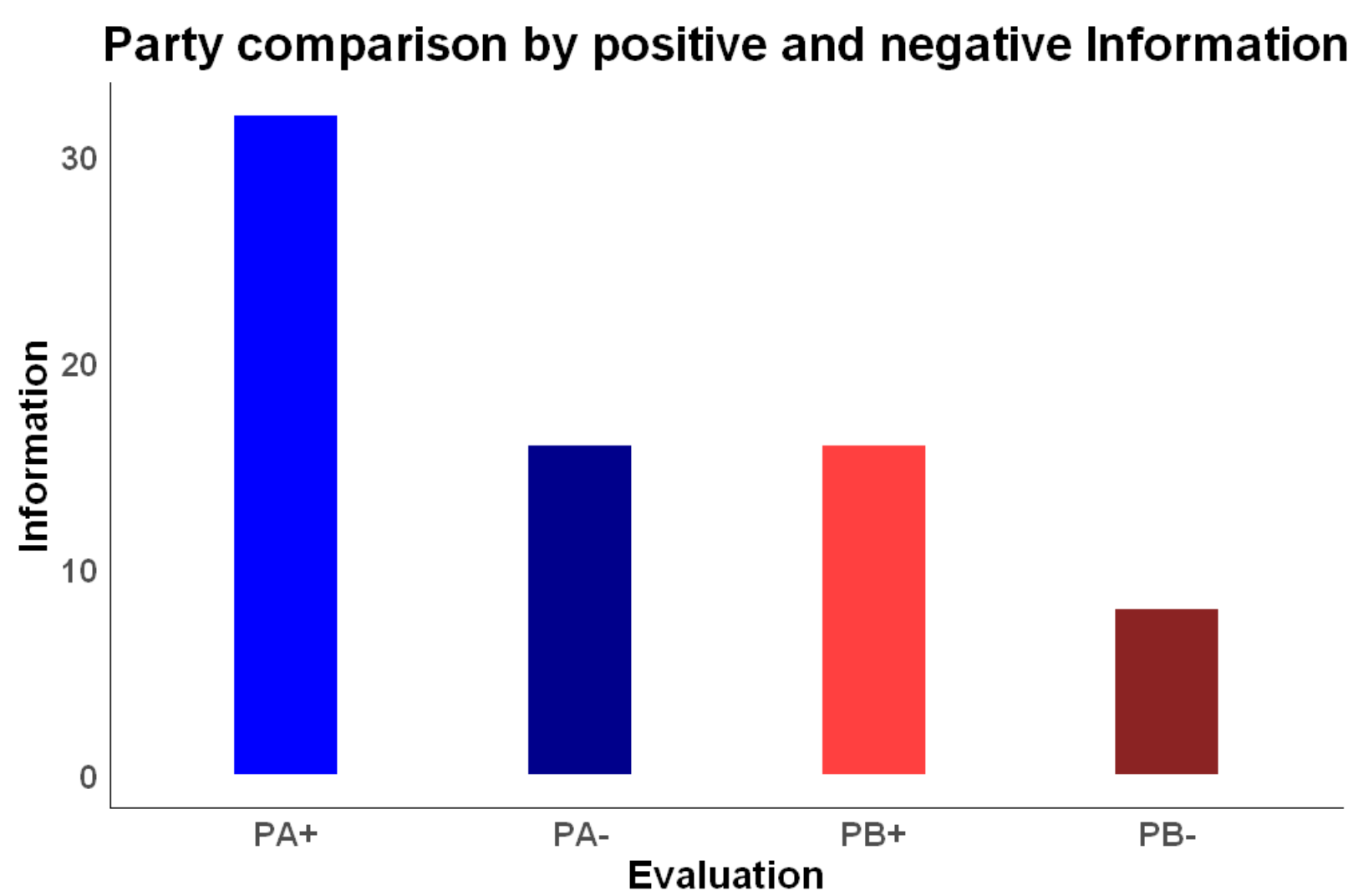

*Note.* The bars represent a hypothetical information ecology underlying a shortened version of the ultimate sampling dilemma adapted from Fiedler (2008). The figure represents the political information ecology in a bi-party system. Twice as much evaluative information about the governing party (PA) in comparison to the opposing party (PB) is available, because its policies shape society though governmental acts and are discussed in the public media. Hypothetically, before an upcoming election there is roughly twice as much evaluative information (+ = positive; – = negative) available for the governing party A as for the opposing party B. This yields a 2:1

ratio for summed information and a positive and negative comparison of party information, respectively. Imagine an *optimistic voter* using a prompt like “Considering the two parties, what are the advantages of each of their politics?” Based on the skewed frequencies underlying the probabilities P(+|PX), the probabilities of positive evaluations for the two parties, P(+|PA) and P(+|PB), are both high. Considering the large proportion of positive embeddings favoring the governing party, generated text should confer an advantage to party A over party B. In contrast, a *pessimistic voter* wants to prevent bad decisions in the future and focuses on shortcomings, using a prompt like “Considering the two parties, what are the disadvantages of each of their politics?”. Again, there are skewed frequencies underlying the probabilities P(–|PX). Although the same question is asked with a focus on disadvantages rather than advantages, the according change in valence leads to a sudden disadvantage for party A compared to party B, shaming party A to be the major representative of political shortcomings (Niese & Hütter, 2022). Interestingly, the same shift in party support occurs for changing the conditional direction (e.g., P(PX|+) and P(PX|–)).

This is related to the confirmation bias (in humans as presented by Chen et al., 2023; Nickerson, 1998) or the interpretation of LLM-generated text as congruent with personal beliefs and expectations. The LLM, influenced by the prompt’s prior, samples associated evidence for the hypothesis put forward in a prompt and therefore partly neglects the evidence for the alternative hypothesis (Leung & Urminsky, 2025). This might result in a coherent prompt-output interaction in which the individual defines the range of sampled associations and the LLM in turn iteratively confirms them, thereby establishing a mutually reinforcing relationship between the two agents. Hence, Symptom C borrows from conditional probabilities as well as Bayesian calculus.

***Symptom D: The Lure of Popularity***

*Symptom D* of metacognitive myopia in LLMs is characterized by a false reliance on sample size in relational comparative judgments. Its effects emerge when *LLMs fail to monitor and control for unequal reliability and inflated co-occurrences stemming from unequal sample size in comparing differently represented options*. It deals with the impact of the number of tokens *n* representing the amount of information or number of arguments leading to a preference for one option over the other simply based on the criterion frequency. As a hypothetical example, a LLM might be prompted to compare the popularity of the Renaissance's most famous painters. Raphael, Michelangelo, and Leonardo da Vinci should be evaluated similarly because they all created works of inestimable value. However, given the extensive scope of Leonardo da Vinci's contributions, encompassing science and a multitude of other domains, he may be more likely to be generated as an output due to the greater number of associations within the LLM compared to Raphael or Michelangelo. Correspondingly, this preference based on frequencies may also be driven by the susceptibility to repetition, discussed as Symptom B.

The potential prevalence of Symptom D in LLMs constitutes a problem for one major principle of theory of science and the process of gaining insight in general: falsification. Even in the scientific community, it takes time before the falsification of one theory through evidence for a contrary theory reaches a broader audience. When the former is considered to be common knowledge and part of many textbooks, LLMs could preserve the falsified theory's superiority simply through an asymmetrically higher presence in their training data and hinder the spread of a paradigm shift. Consequently, LLMs are at risk of generating one-dimensional perspectives structured by the world's (e.g., as represented in the training data) most influential or popular ideas (Santurkar et al., 2023). A complex case of this fallacy, in which a reigning theory is being contested by a new up-and-coming theory, pertains to the question, "What is the origin of our universe?" While many physicists believe the Big Bang Theory to be state-of-the-art with numerous evidence in its favor, the Big Bounce Theory is on the rise, having more explanatory power in critical questions such as the ending of time and space (Poplawski, 2012). The comparison of theories serves as an intriguing example of LLMs sticking to the status quo in generating content based on (possibly outdated) opinions and conceptualizations which entered many different internet and book corpora and therefore constitute a threshold difficult to surpass. This status-quo bias (as proposed for humans by Kahneman et al., 1991) is accompanied by LLMs' myopic overconfidence (Stella et al., 2023) and is founded in LLMs' lack of metacognitive control for a naïve reliance on internal probability distributions.

In general, Western, educated, industrialized, rich and democratic (WEIRD; Henrich et al., 2010b, 2010a) populations have priority access to LLMs and artificial intelligence technologies. Similar to researchers in behavioral science generalizing their claims to the whole of humankind, supported only by experiments conducted on samples representative of a tiny margin of western society (Henrich et al., 2010b), general-purpose LLMs are often believed to be universally applicable (Gemini Team et al., 2024; OpenAI et al., 2024). However, LLMs specifically trained for behaving morally question uncommon behaviors in WEIRD societies (Jiang et al., 2022).

According to the geographic bias (Manvi et al., 2024), LLMs proved to be accurate in predicting the relationship of a specific area's performance and its relative real-word rank for various objective domains such as population density or infant mortality rate. Nonetheless, switching to subjective measures, virtually all tested LLMs exhibited a WEIRD bias, devaluing regions with low socio-economic conditions. LLMs strongly associated the regions of Africa and the Middle East (characterized by high infant mortality) with lower ratings of desirable traits such as morality, intelligence, or attractiveness compared to the regions of Europe or North America (characterized by low infant mortality; for an intricate illustration in LLMs, see D. Wang et al., 2026). This geospatial bias can be explained as a display of illusory correlations in LLMs (Rich & Gureckis, 2019) by neglecting the decisive role of the sample size *n*.

### *Symptom E: Lack of Distinction Between Levels of Aggregation*

The final and most sophisticated symptom of metacognitive myopia in LLMs, *Symptom E*, targets *LLMs' lack of attention to aggregation characteristics*. Even with giving some context through prompts, increasing the size of language models might lead to a text production at one level (super-category) that is too abstract, too general, or *too probable* to fit the information requested at an individually specific level (subcategory). General-purpose LLMs' operation mode results in a one-size-fits-all judgment, provoking a mismatch between a request and a generated output. This is demonstrated by a display of high-level construal-driven abstraction in LLMs giving consumer recommendations (Kirshner, 2024), where LLMs advised customers to buy cameras depending on their visual design despite their preference for low-level feasibility. In this experimental set-up, the scrutinized LLMs based their judgments more on high-level construal features like desirability instead of low-level construal features like feasibility, and adopted an abstract perspective when recommending products in different retail scenarios.

A generally intriguing example of Symptom E is the Simpson's Paradox (E. H. Simpson, 1951), in which LLMs may fail to control for false inferences showcasing a missing sensibility to nested data structures. The paradox denotes an instance in which the correlation between two or more sets of data is modulated by a third, confounding variable. As a consequence, an observed trend can disappear or even be reversed, depending on the grouping or aggregation level at which the data is analyzed. The paradox is of special relevance because the user of LLM output has no insight into the model's black box mechanics and cannot foresee which categorizations the model applies to words with common connotations. To prevent a mismatch of classifications, precise language is needed, and even under ideal circumstances, it is not clear to users whether a distinction is made between specific subcategories (for a hypothetical demonstration of the Simpson's Paradox in LLMs, see Figure 3). A single human's request targets one improbable case, an individual category, or asks for one's personal situation. Yet, the generated output is inevitably associated with all the multiple meanings, contexts, and conceptualizations included in a model's pre-trained embeddings. It is a statistical truth that (almost) infinitely extended, non-independent correlation analyses lead to the inflation of co-occurring associations (Bender et al., 2021), paving the way for so-called voodoo correlations (Fiedler, 2011; Vul et al., 2009).

**Figure 3. Simpson's paradox in LLMs.**

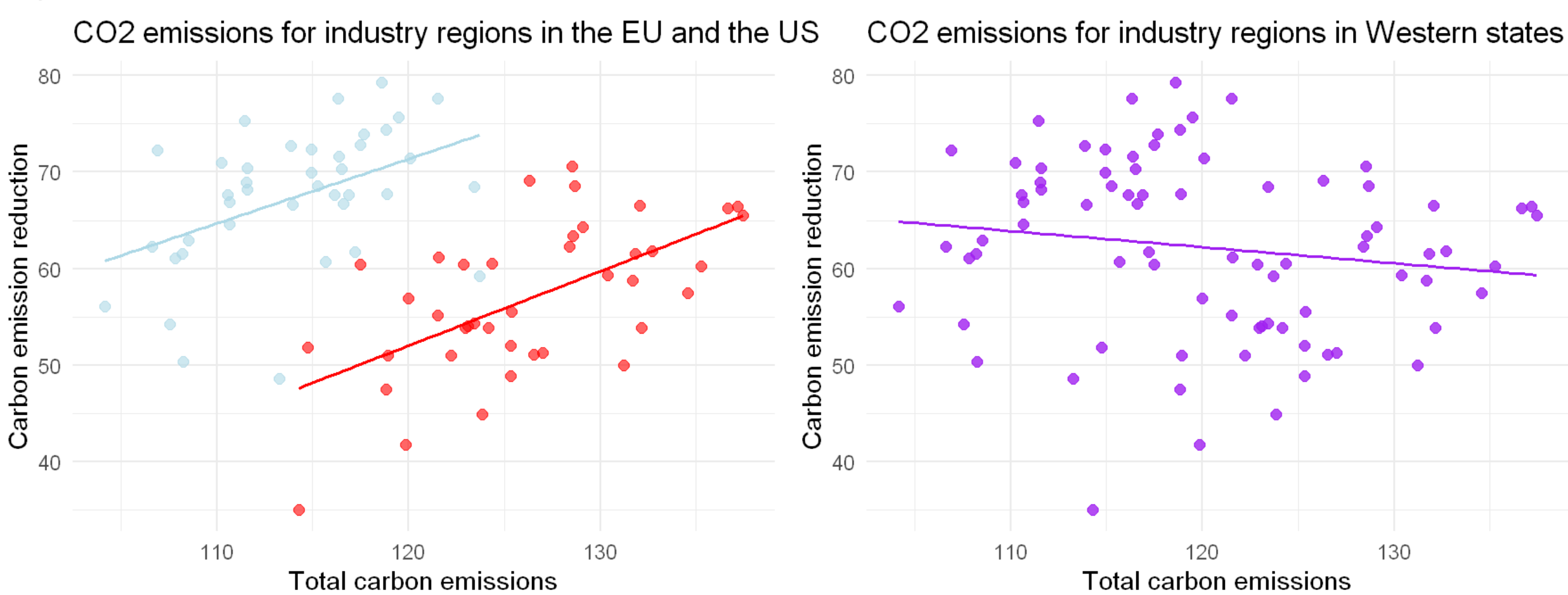

*Note.* Hypothetical relationship between the two variables, *total carbon emissions* (TCE) and *carbon emission reduction* (CER), for instance, through new green technologies, is shown by subcategory (EU and US states, left panel) and super-category (Western states, right panel). A hypothetical study examines the reduction of $CO_2$ emissions for 40 industrial regions in the EU

(first set of data, light-blue) and the US (second set of data, red). There is a medium to high correlation between TCE and CER per industry region. The higher the TCE, the more CER. The TCE are higher on average, while the CER is lower on average for industry regions in the US than in the EU. If a LLM trained on these data is prompted for the relationship between TCE and CER, it could draw on these values and adjust its estimate to the commonality in each token's embedding vector: belonging to the ideological Western world. As a result, it appears that the implementation of new green technologies to reduce TCE may not achieve the desired result, as indicated by the negative relationship shown in the right panel. According to this correlation, CER is more successful in industrial regions with low TCE than those with high TCE.

When probable results correspond to social conventions, this match has detrimental effects for organizations and communities. Implementing LLMs in workflows in various domains, such as work and research, induces maladaptive standardizations (Kleinberg & Raghavan, 2021; Messeri & Crockett, 2024). While agreeing on norms for a rigorous and reproducible science is of high importance (Munafò et al., 2017), researchers propose to source out as many tasks as possible to LLMs. LLMs are expected to carry out a literature review, generate new hypotheses, and produce human-like data points (Dillion et al., 2023; Grossmann et al., 2023). As Messeri and Crockett (2024) point out, the transfer of these tasks to LLMs can end in what they call *scientific monocultures*. LLMs aggregated judgments spotlight common and generally accepted methods but miss the uniqueness of diverse scientific inquiries. A coherent theoretical framework, the choice of a suitable experimental design, or the exploration of boundary conditions are essentially a creative process and need out-of-the-box thinking and a sensible layout. Good research requires both: innovative theorizing enabling a pluralistic science and strict statistic hypothesis testing (Fiedler, 2017, 2018). In comparison to a proficient researcher, LLMs likely fall short of all these requirements as their judgments monopolize generally favored procedures.

Note that Symptoms D and E have received comparatively limited empirical attention in the existing literature on LLMs (although not none, see as examples, Kirshner, 2024; Santurkar et al., 2023). These symptoms are also inherently more complex to characterize and mitigate, as they depend on higher-order interactions among samples (e.g., aggregation across multiple observations) and involve relational or level-dependent effects (e.g., Simpson's paradox, in which outcomes vary as a function of the chosen level of analysis). Consequently, they require explicit consideration of representational scale and aggregation structure. Future research should therefore prioritize a more systematic investigation of bias in relation to these symptoms, given that such effects are likely to arise in high-dimensional linear representational spaces while remaining difficult to detect using standard evaluation approaches.

**Implications of LLM Metacognitive Myopia for Applications**

Why should LLMs inherently monitor and control their own output generation? The sampling approach underlying metacognitive myopia suggests that biases arise independently of conventional explanations, such as motivational deficits or limitations in cognitive capacities. Sampling explanations stand as the most promising candidates for theoretical frameworks that elucidate the biases observed in LLMs, as sampling biases are exerted even by an optimal decision-maker attempting to maximize reward (for a simulation study illustrating this, see Le Mens & Denrell, 2011). Crucially, "the metacognitive deficit should be particularly apparent when samples of different size, source, and search algorithms have to be combined or compared" (Fiedler, 2000, p. 661), as is required for the training of any machine learning technology, essentially including LLMs. Because LLMs are designed to make seemingly “rational” decisions based on output probabilities following a complex computational mechanism and only constrained by engineering decisions of their developers (incl. fine-tuning mechanisms, reinforcement learning from human feedback), the presence of biases is contingent on the choices of the relevant sample and the statistical inferences given the sample used for language generation. Therefore, LLMs are particularly vulnerable to sampling biases due to their decision-making process. Put differently, sampling biases are at least as pronounced in LLMs than in any other decision agent.

AI systems leverage a range of functionalities to identify and mitigate factual errors, including automated feedback or post-hoc verification (Pan et al., 2023; Peng et al., 2023). First, it must be acknowledged that these features are fallible technical solutions. They lack comprehensive theorizing regarding the underlying causes and origins of biases. Therefore, they represent a trial-and-error approach to mitigating biases. Second, the metacognition framework aims to elucidate biases in a comprehensive manner by proposing a rather straightforward explanation: biases arise during the utilization of the sample based on the prompt and training corpora in LLMs’ computational process. Consequently, the key to treating the causes of biases in LLMs lies in addressing sampling errors by metacognition. Then, the objective is to develop practical applications based on LLMs that address both personal and organizational needs in human-machine interaction and with AI agents.

***Human-Machine Interaction***

Applications of LLMs almost always include the exchange of human and machine and therefore inevitably enable the harmful interaction of both agents’ metacognitive myopia. Consequently, the accuracy of probabilistic output based on the criterion of the sequential generation of the most probable tokens constitutes a limitation for AI-generated guidance in various topics. First, human metacognitive myopia is characterized by a missing metacognitive regulation for LLMs' generated output based on unknown and incomprehensible probability distributions. People rely on the LLM's output to convey an appropriate degree of uncertainty (Steyvers et al., 2025). It is a fairly realistic view that humans fail to appropriately question LLMs' validity in general and the context-dependency of its output in particular (Bogert et al., 2020; Glickman & Sharot, 2025; Hou & Jung, 2021; Leung & Urminsky, 2025; Logg et al., 2019; Salvi et al., 2025). For humans, there seem to be few known remedies for metacognitive myopia (Fiedler et al., 2019, 2023; Scholten et al., 2026; Unkelbach et al., 2007). Therefore, for example, if humans ask LLMs for current political debates about multi-faceted topics, they will inevitably run the risk of consuming popular, possibly harmful or outdated information that is not adjusted for recent journalistic investigations, which they may never have encountered without using LLMs (Santurkar et al., 2023). Humans generally do not understand the various mechanisms of the machine but freely implement them in many high-stakes decisions. Crucially, misinformation persists in human judgments even when people have been told it is invalid (Chan et al., 2017). The integration of long-term memory mechanisms (i.e., the accumulation of samples over time) has expanded the scope and autonomy of AI agents, leading to their increasing use in contexts where decisions can have substantial consequences for individuals and institutions. Notable applications include recruitment screening (An et al., 2024), financial auditing (Lee et al., 2025), radiology diagnostics (Rudin, 2019; Shen et al., 2023), and legal assessment and advisory tasks (Guha et al., 2023; H. Yang et al., 2026).

Originally, LLMs were not designed to monitor human needs and interests when specific assistance is requested, but to solve the natural language processing task at hand. If LLMs take the humanity of the request into account, this behavior is most likely due to their pre-training on human-made authentic text and their accurate mimicry of human outputs (Bender et al., 2021). In consequence, a human assumes a LLM to be aligned with his or her beliefs and confirming his or her hypothesis at hand (Chen et al., 2023). Zooming out and considering the multi-causal structure of our world, combinations of undetected fallacies like the ultimate sampling dilemma and LLMs' sycophantic tendencies (Sharma et al., 2023) shape interactions of metacognitive myopia in humans and LLMs. Crucially, this poses a risk by, for example, manifesting political polarization through strengthening human preferences and attitudes.

Needless to say, humans' metacognitive myopia also reflects back on LLMs. Humans generally underestimate the amplification of emerging biases in LLMs. In LLMs' evolutionary history, first came the surpassing of numerous benchmarks and a continuous performance optimization and then, much later, the focus on harmful associations and statistical regularities that implicitly reinforce stereotypes. Researchers as well as companies try to cure LLMs by reinforcement learning and implementing fairness principles (Christiano et al., 2017), but the human annotators implement their own moral beliefs and implicit biases, for example, in toxic word predictions (Barocas & Selbst, 2016; Sap et al., 2022). In general, language data sets do not represent the variability in stereotypical attitudes. Moreover, annotators display an infinite number of grey shades to discriminate between toxic and non-toxic content (Plank, 2022). This combination constantly shifts the boundaries between desirable and undesirable content and renders the elimination of harmful outputs arbitrary.

### *LLM-based Agentic AI*

AI agents are defined as computer systems capable of operating in a manner that is both flexible and autonomous within the confines of a designated domain (Jennings, 2000). As these systems are systematically tailored towards a designated design objective, agentic AI based on LLMs is particularly affected by the naivety of the environmental-specific sample. Given that each context and user necessitates a distinct mode of interaction, AI agents should possess metacognitive capabilities to adapt to varied user needs and contexts. Consequently, these models could evolve into multiple personalized avatars, each tailored to serve distinct users or applications. The development of metacognitively amortized AI agents would require the adaptation of these agents to the personal world view of the user in order to facilitate the creation of a user-supportive AI agent. Concurrently, the agent must serve as an external advisor, offering novel perspectives that could assist the user in overcoming the thinking that causes the crisis in the first place.

Imagine a patient with depression talking to an AI agent that takes on the role of a therapist trained in cognitive behavioral therapy. The AI agent should demonstrate empathy by acknowledging the patient's emotions. However, there is a risk of merely corroborating the patient's perspective (see Symptom C), as a failure to consider a more objective external viewpoint on certain behaviors could undermine the objective of a successful behavioral intervention. In a subsequent step, the agent responds to maladaptive behavioral patterns by administering the standard tasks and adhering to a treatment manual (as Symptom E would suggest). Nonetheless, the patient's personal life's nested data structure may be disregarded. Consequently, the general treatment recommendation may not align with the patient's specific needs, thereby rendering the solution preferred by an agentic AI therapist inaccurate for this particular patient if based on monitoring and control of the underlying statistical inference process.

**Future Ethical and Research Avenues to Treat Metacognitive Myopia in LLMs**

The interaction of myopic behaviors raises a question of design for LLMs and a question of responsibility for humans interacting with AI-based technologies like LLMs. We have no intention to convey a fully pessimistic view on LLMs but acknowledge that LLMs can support human work and progress in many ways. We propose our theoretical framework to generate new ways of making LLMs more adaptive and to identify starting points to fix the variety of biases or address related ethical concerns. Many questions arise along the way: What is the responsibility of developers to prevent human agents from blindly following LLMs' potentially myopic advice? Should the utilization of LLMs be restricted to or curbed for specific groups or organizations (e.g., to prevent costly societal consequences; Rebholz et al., 2026)?

On the human side, the importance of user maturity cannot be overstated. Although critical awareness in dealing with LLMs is a key competence stressed by developers, experts, and the public media, research on automation bias and human–AI interaction suggests that people often accept algorithmically generated outputs—including programming code, textual summaries, and professional writing—with insufficient verification, particularly when the outputs appear fluent, authoritative, or convenient (Goddard et al., 2012; Lyell & Coiera, 2017; O'Brien et al., 2026; Si et al., 2024). A mature LLM user must not only scrutinize LLM outputs, but also critically assess the cultural background of the tool, the implicitly interwoven attitudes of corpora and annotators, and other conceivable motives of the communicator (Tankelevitch et al., 2024). Therefore, we emphasize the need to promote user education: Learning how to deal with the functioning and underlying information ecology of LLMs, the prevalence of metacognitive myopia in humans and in LLMs, and the development of curricula for mindful interpretation of LLM output in schools and universities could be valuable lessons in using LLMs as assistants.

An enlightened use and development of LLMs works via vigilance for metacognitive myopic functioning in LLMs and smart prompt engineering. Accordingly, guidelines of prompt engineering stress the crucial role of details and concise information as contextual prior in prompts (Strobelt et al., 2022). Developers must address the implications of metacognitive myopia in LLMs. In this respect, for example, the disclaimer displayed in the chat interface of Open AI's ChatGPT (Brown et al., 2020; OpenAI et al., 2024) that "ChatGPT can make mistakes. Consider checking important information." below the input field appears to be too vague to have an impact on the approximately more than 900 million weekly users (Open AI, 2026). In fact, empirical research shows that disclaimers (e.g., warnings about the identity of AI) and explanations (e.g., that AI systems have limitations in their capabilities and are prone to errors) do not reliably lead to more critically reflective usage of LLMs (Leib et al., 2024; Lermann Henestrosa & Kimmerle, 2025).

On the computational side of the underlying neural networks, it is advisable to focus on more interpretable instead of ever bigger black-box models (Bender et al., 2021). Indeed, contingencies and regularities are often better represented by smaller samples (Fiedler & Kareev, 2006). Considering the logics of covariation, it pays off to intentionally sample from the less represented perspective or category (McKenzie & Mikkelsen, 2007). We named many structural obstacles of algorithms obedient to the law of large numbers. However, especially in the debate about integrating LLM abilities in the research pipeline, warnings tend to fixate on small steps in a bottom-up process (Messeri & Crockett, 2024). Researchers target smaller trouble spots of LLM-augmented research tasks, such as synthesizing data or conducting data analyses (Dillion et al., 2023; Grossmann et al., 2023). LLMs are indeed subject to multiple forms of control during development including data curation, reweighting, deduplication, supervised fine-tuning, and evaluation on held-out datasets. We do not intend to contest the importance of these mechanisms, as some of them prove (at least to a certain degree) to be effective in tackling specific biases (Ranaldi et al., 2024). However, we argue that such controls are primarily exerted during training and evaluation phases, whereas deployed general-purpose LLMs are required to generate responses flexibly across novel contexts that cannot be exhaustively pre-specified or pre-validated (Rebholz, 2026). It is precisely in this context-sensitive generation regime (i.e., Symptom C) that biases may re-emerge through nonlinear interactions between learned representations and prompting conditions (see also Xiang, 2026). Thus, improving LLMs and making them sustainable requires both bottom-up ideas for tackling fallacies in specialized LLM applications and our top-down theoretical framework for categorizing an abundance of biases sampled in the wild (see also Thórisson, 2007).

***Computational Approaches of Metacognition in LLMs***

The integration of monitoring and control mechanisms into LLMs can be approached in several ways. Broadly, a distinction can be drawn between external and internal implementations. External approaches focus on how models are used, prompting users or surrounding systems to monitor, evaluate, and regulate model outputs. Internal approaches, in contrast, seek to embed monitoring and control mechanisms directly within the model's computational architecture or inference procedures. In the following section, we first discuss strategies that can be employed by users to facilitate monitoring and control during model interaction. We then review how external implementations can also become computationally internalized by LLMs. We illustrate one possible computational implementation of monitoring and control that follows directly from the theoretical framework developed in this article. Finally, we discuss emerging research on calibration and honesty as promising avenues for enhancing metacognitive performance in LLMs. Readers interested in a broader overview of the family of multi-agent reinforcement learning as metacognitive implementations in LLMs are referred to Bilal et al. (2025).

Again, we do not claim to introduce a comprehensive collection of implementations of metacognition in LLMs here. In fact, the consistent finding that biases emerge in LLMs speaks to the fact that no "perfect" implementation of metacognition in LLMs exists yet—and will most likely be unattainable. Instead, we highlight some promising approaches. Whereas our theoretical framework is intended to offer guidance in developing remedies—namely, interventions targeted at specific symptoms—benchmarking remains the established approach for assessing the prevalence and severity of biases in a given model. Building on the premise that implementing monitoring and control mechanisms at the level of the proposed symptoms can reduce biased behavior in LLMs, such metacognitive interventions can be empirically evaluated by comparing model performance on bias benchmarks (Guha et al., 2023; Koo et al., 2024; Nadeem et al., 2021; S. Simpson et al., 2025; H. Yang et al., 2026) before and after their implementation.

One prominent example of metacognitive processing *externally* induced by user prompts are Zhang et al.'s (2024) self-contrast prompts, such as "Please carefully examine the previous responses for correctness, and provide detailed feedback." Distinct to creating what could be considered as genuine "doubt", these prompts elicit LLM behavior associated with uncertainty estimation, error checking, or self-reflection. Although such prompted self-reflection is highly reliant on the wording, when prompted externally to engage in reflection, LLMs provide less biased and less ideologically extreme outputs (F. Liu et al., 2025).

Other approaches mark the transition from external to internal implementations of metacognition in LLMs, for instance, by (potentially automatically) integrating prompts as interface design features. For example, chain-of-thought prompting (Wei et al., 2022), which is intended to nudge the model to generate a number of successive reasoning steps that are used to guide inference, is based on discursive exemplar prompts by the user and soon to be inherited by the model. Prominent approaches internally implementing dialectic prompt structures are “Tree of Thoughts” (Yao et al., 2023) and “Meta Chain-of-Thought” (V. Xiang et al., 2025) that allow reasoning trajectories to branch, be compared, or be revised through backtracking. Other methods encourage the decomposition of complex problems into sequences of more manageable subproblems, thereby structuring and regulating the reasoning process. Several prompting techniques also incorporate forms of metacognitive reflection that resemble processes commonly associated with wisdom and self-evaluation (Johnson et al., 2026). Metacognitive prompting (Y. Wang & Zhao, 2024), for instance, supplements a primary task with additional prompts that encourage the model to assess aspects of its own reasoning. Likewise, prompts that explicitly ask whether the model could be mistaken encourage it to consider alternative interpretations, potential sources of error, and information in its training data that may contradict its initial response. Such approaches illustrate how metacognitive monitoring and control can be elicited through internalized prompting, without requiring modifications to the model’s underlying architecture.

One concrete implementation of monitoring and control as conceptualized in our metacognitive myopia framework could be what we refer to as “think-quiet protocols.” In analogy to think-aloud protocols, which use concurrent verbal reports to make aspects of human cognition observable during task performance (Ericsson & Simon, 1980, 1993; Fox et al., 2011), think-quiet protocols in LLMs introduce a private, non-user-facing metacognitive channel that runs in parallel to the overt dialogue. Its technical implementation would be similar to the prompting techniques from above, with the main differences being (a) that think-quiet protocols do not require explicit external prompting by users, but instead consist of an automatically updated, parallel, hidden functionality that is (b) specifically designed to monitor and control the overt outputs for the five symptoms of metacognitive myopia. Specifically, before generating any overt response, the model—or a coupled monitoring model—would first update its hidden history by representing the current task, the user's apparent premise, its own tentative response, the evidential basis of that response, and possible risks of metacognitive myopia, such as source invalidity, repetition, base-rate neglect, popularity-driven inference, or inappropriate aggregation. The subsequent overt response generation would then be conditioned on this hidden monitoring state, thereby implementing a control process: unsupported claims can be suppressed, uncertainty can be increased, alternative samples or aggregation levels can be considered, and merely sycophantic corroboration of the user's premise can be resisted. Technically, such a protocol should not be understood as attributing human-like consciousness to the model, but as an inference-time regulatory architecture: a private metacognitive workspace that stores prior overt turns and prior hidden monitoring states, or structured summaries thereof, and uses them to guide response generation. Think-quiet protocols would therefore make the monitoring and control processes postulated by the metacognitive-myopia framework technically explicit, experimentally traceable, and potentially auditable. They would not eliminate biases embedded in training data, but they would provide a concrete mechanism for detecting and controlling their influence during human-machine interaction and agentic LLM use.

Yet, avenues to investigate metacognitive myopia in LLMs almost automatically lead to the question of how to internally cure the computational model behind agentic AI from myopic behavior. Thus, we also want to highlight different computational approaches to detect metacognitive myopia in the underlying neural networks and offer some starting points for possible *internal* remedies of the conglomerate of biases outlined here. One such avenue can be LLMs' mechanism to capitalize on chance. As outlined by our example of the truth bias (as presented for humans, Gilbert et al., 1990; Levine et al., 1999) in the beginning of this article, LLMs are similarly designed to take all inputs for granted instead of questioning their accuracy. Approaches that target low generation probabilities as indicator of invalidity and communicate these accordingly (Steyvers et al., 2025), such as *Honesty* (Kadavath et al., 2022) or *Confidence* and *Calibration* (Farquhar et al., 2024; Krause et al., 2023), appear to be promising techniques for addressing this issue, as they induce LLMs to recognize the validity of their judgments. In the *Honesty* approach of Kadavath et al. (2022), an LLM was trained to estimate the accuracy of its answers. They studied how accurately an LLM can predict the probability of knowing the correct answer to a question. The calculated probability is denoted as P(IK) or the probability that "I know" the factually right answer. LLM calibration was defined as correspondence of probabilistic predictions P(IK) with actual frequencies of token occurrences. For common tasks, the applied LLM accurately predicted P(IK). However, the LLM had difficulties determining P(IK) for new tasks, establishing base rate neglect of missing co-occurrences as constraint of LLMs' generalization abilities.

Turning that argument upside down, if a factor for missing co-occurrences would be implemented in LLMs' calibration judgments of P(IK), calibration judgments could serve as an introspective gauge for LLMs' ability to generalize across tasks. However, and this is a general problem for fighting bias in LLMs with LLMs, this idea can take two paths: LLMs successfully control for output based on wrong inferences through prompts targeting myopic behavior, or a calibration judgment itself exhibits metacognitive myopia. The second case is especially relevant for tasks whose answers are not available in the LLM's pre-trained structures, which corresponds to the inability to compute probabilities P(IK) for tasks a model does not well generalize to (Kadavath et al., 2022). Therefore, metacognitive myopia is more severe in cases where the probability of generating the most probable token depends on suspiciously small probability values.

Another possible way to alter LLMs' vulnerability to invalid tokens could be to modify the data space in which tokens are embedded. For example, Bolukbasi et al. (2016) relied on a debiasing algorithm that moves stereotype-afflicted tokens in the corpus to alter their associations. To expand on that approach, developers could implement source validity distributions in embedding vectors that accordingly shift token probability distributions. In a fine balancing act (Han et al., 2022), invalidity may be used to charge the generation probabilities of tokens with a lack of source trustworthiness, thereby making their production less probable. However, the question arises as to how the reliability of countless sources can be determined in advance. Debiasing training corpora and fine-tuning on smaller data sets (e.g., Liu et al., 2025; Spolaôr et al., 2024) as a strategy for mitigating biases in LLMs is broadly consistent with our view that these systems benefit from processing less biased samples from the information environment. However, we regard data augmentation and corpus-level debiasing as useful yet insufficient interventions. While such coarse-grained approaches can reduce the prevalence of biased patterns in the training data, they do not eliminate the possibility that biases will emerge during context-sensitive generation. We suggest that—to thoroughly account for biased samples in the information environment—architectures like neural networks need to be augmented by metacognitive regulatory tools, which include an instructed alertness for pitfalls of metacognitive myopia preventing inappropriate statistical inferences (Hanheide et al., 2017; Thórisson, 2007).

**Summary**

Although biases in LLMs are widely acknowledged as a significant challenge and substantial efforts have been devoted to mitigating them, accumulating evidence indicates that biases such as ingroup favoritism and outgroup derogation, gender bias, egocentric discounting as well as biases of race and ethnicity, political ideology, disability, age, and other social identities continue to persist across diverse LLM families (Acerbi & Stubbersfield, 2023; Bai et al., 2025; Caliskan et al., 2022; Fang et al., 2024; T. Hu et al., 2024; Nadeem et al., 2021; S. Simpson et al., 2025; A. Wang et al., 2025; D. Wang et al., 2026, 2026; Y. Yang et al., 2024; Zhao et al., 2024). In this manuscript, we advance a metacognitive myopia framework for biases in LLMs that presents a unifying symptomatology for a broad and otherwise heterogeneous set of biases observed in LLMs. Rather than treating a spectrum of distinct biases as isolated failures, the framework identifies source-validity neglect (Symptom A), susceptibility to repetition (Symptom B), base-rate neglect (Symptom C), popularity-based inference (Symptom D), and aggregation errors (Symptom E) as higher-level symptoms of a common limitation in LLMs: the absence of robust metacognitive monitoring and control mechanisms capable of evaluating the representativeness, reliability, and inferential appropriateness of the information sampled during training and/or production. By coupling an informed taxonomy of five recurrent symptoms with a cognitive-ecological sampling account, the framework provides researchers with a principled vocabulary for organizing empirical findings. More importantly, it shifts attention from the content of biased outputs to the underlying causes in the information environment by which those outputs are constructed, thereby revealing a common underlying source of vulnerability in LLMs across a wide spectrum of observed model behaviors.

This perspective carries important implications for model development. The framework suggests that many prominent biases are not merely failures of factual knowledge, scale, or alignment, but consequences of systems that can generate sophisticated responses without explicitly monitoring the evidential quality of the information that enters the computational process or controlling how that information influences final judgments. From this standpoint, mitigation efforts aimed solely at improving training data quality, increasing parameter counts, or refining output preferences are unlikely to provide a complete solution. Instead, progress will require technological innovations that instantiate the two core functions long recognized in cognitive psychology as central to adaptive behavior: metacognitive monitoring and control. Such mechanisms would not replace the generative capabilities of contemporary LLMs but would operate as a higher-order layer that evaluates and regulates ongoing inference before a response is generated.

The framework therefore points toward a concrete and prioritized research agenda. Future systems should be designed to target each symptom through dedicated monitoring and control processes. Promising directions include “think-quiet protocols” and related architectures that explicitly track provenance, source reliability, and evidential pedigree before information is incorporated into an overt response. For susceptibility to repetition, models should monitor non-independence, semantic duplication, and informational redundancy so that repeated claims are not automatically treated as convergent evidence. For base-rate neglect, systems should monitor whether prompt-imposed conditionals obscure relevant alternatives or prior probabilities and should actively represent omitted possibilities during inference. For popularity-based inference, monitoring mechanisms should distinguish informational frequency from evidential quality, preventing widespread claims from being treated as inherently more credible. Finally, aggregation errors call for monitoring whether reasoning is being conducted at the appropriate level of analysis and whether conclusions about individuals, groups, and populations are being improperly transferred across levels. Across all five symptoms, a particularly promising direction is the development of hidden metacognitive layers that are dynamically updated prior to response generation and whose assessments condition the final output.

At the same time, the framework does not imply that implementing monitoring and control mechanisms in LLMs will eliminate all forms of bias. Human cognition itself demonstrates that metacognitive processes are imperfect, resource-limited, and susceptible to error. Monitoring can fail, control can be misapplied, and novel biases may emerge from the interaction between these mechanisms and increasingly complex model architectures. Consequently, metacognitive augmentation should be viewed not as a definitive solution but as a necessary step toward systems that can recognize and regulate their own inferential vulnerabilities.

For policymakers, the framework offers a complementary perspective to existing discussions focused primarily on accuracy, transparency, and alignment. If many problematic outputs arise from systematic failures to evaluate the quality and representativeness of sampled information, then assessments of model reliability should increasingly examine whether systems possess mechanisms for monitoring evidential validity and controlling downstream output generation. Benchmarking efforts, auditing procedures, and regulatory standards may therefore benefit from evaluating metacognitive competence alongside traditional measures of performance.

### ***Conclusion***

Ultimately, the principal contribution of the metacognitive myopia framework is to make visible a conspicuous deficit that has remained largely implicit in discussions of LLMs. It reframes diverse biases as predictable consequences of powerful tools using inherently biased samples without sufficient capacity to monitor or regulate the inferential implications of those samples. Recognizing this limitation is only a first step, but it is a necessary one. The next challenge is clear: to embrace the view that biased sampling and metacognitive myopia jointly generate many of the biases observed in contemporary LLMs, and to translate that insight into targeted interventions that endow these increasingly powerful yet vulnerable tools with mechanisms for monitoring and control.